\ifdefined\pdfoutput\pdfoutput=1\fi % arXiv: "\pdfoutput=1" within the first 5 lines makes AutoTeX run pdflatex (figures are PDF)
\ifdefined\pdfminorversion\pdfminorversion=7\fi % figure files are PDF 1.7; avoids pdfTeX's "PDF inclusion: found PDF version" warning
\documentclass[11pt,a4paper]{article}

\usepackage[utf8]{inputenc}
\usepackage[T1]{fontenc}
\usepackage{microtype}

\usepackage[a4paper,margin=1in]{geometry}

\usepackage{amsmath}
\usepackage{amssymb}

\usepackage{booktabs}
\usepackage{graphicx}
\graphicspath{{figures/}}
\usepackage[font=small,labelfont=bf]{caption}

\usepackage{enumitem}

\usepackage{url}
\usepackage[hidelinks]{hyperref}
\hypersetup{
  pdftitle={Self-Supervised Pretraining for PHM: What Transfers, What Does Not, and What Federating It Costs},
  pdfauthor={José Sanz Durán and David Solís Martín},
  pdfkeywords={prognostics and health management; self-supervised learning; time-series foundation models; federated learning; transfer learning; negative results}
}

\newcommand{\sci}[2]{\ensuremath{#1{\cdot}10^{#2}}}
\newcommand{\dmodel}{\ensuremath{d_{\mathrm{model}}}}

\title{\bfseries Self-Supervised Pretraining for PHM:\\
  What Transfers, What Does Not, and What Federating It Costs}
\author{%
  José Sanz-Durán \qquad David Solís-Martín\\[3pt]
  \normalsize Department of Computer Science and Artificial Intelligence\\
  \normalsize Universidad de Sevilla, Seville, Spain\\[3pt]
  \normalsize \texttt{jsandur@outlook.com} \quad \texttt{dsolis@us.es}%
}
\date{}

\begin{document}

%% ===== S01_front: Front matter (\maketitle, abstract, keywords) =====
\maketitle

\begin{abstract}
\noindent Self-supervised pretraining has produced capable time-series foundation models, but its
value for Prognostics and Health Management (PHM) has not been measured under
leakage-controlled conditions. We measure it on a corpus of 47 public PHM datasets, harmonized to a
single tensor contract and partitioned so that 11 datasets are withheld from pretraining in full.
Transfer turns out to be task-dependent: a compact channel-independent Transformer encoder
(801,808 parameters) pretrained with masked patch prediction raises fault-classification
macro-F1 from 0.350 to 0.812 on CWRU and from 0.569 to 0.951 on HSG18, yet on CMAPSS
remaining-useful-life estimation no pretrained checkpoint reliably improves upon a
from-scratch baseline. We pretrain under two regimes---centralized, and simulated cross-silo
federated with ten clients grouped by physical domain---holding architecture, objective, source
data and optimization budget fixed. Two further findings are negative. At matched budget the
centralized encoder leads by 15.5--20.2 macro-F1 points in four of four comparisons, although
the comparison is confounded by a learning-rate schedule that favours the centralized arm and
rests on a single pretraining run per regime; on RUL the two regimes are indistinguishable.
And the intrinsic SSL loss is not a valid selection criterion: the centralized encoder more
than halves the federated reconstruction loss without a corresponding gap on RUL, and a
server-momentum sweep moves downstream validation macro-F1 by up to 0.29 while the SSL
loss varies by 0.021. Transfer also hinges on the backbone learning rate: the same checkpoint
yields 0.169 or 0.829 macro-F1 on CWRU depending on a single order of magnitude. We
release the corpus pipeline, the protocol and the code.
\end{abstract}


%% ===== S02_intro: Section 1 Introduction =====
\section{Introduction}\label{sec:intro}

Industrial maintenance has shifted from corrective and calendar-based strategies towards condition-based and predictive policies. Prognostics and Health Management (PHM) provides the analytical frame for that shift: it turns sensor signals into diagnosis of the current health state and prognosis of the future one, typically instantiated as fault classification, remaining useful life (RUL) estimation and anomaly detection~\cite{huang2024phm,fink2020potential}. The practical bottleneck is well known. Raw condition-monitoring signals are abundant; labelled failures are scarce, expensive and often ambiguous across organizations. That asymmetry---abundant unlabelled signals, scarce labels, high heterogeneity---is precisely the setting in which self-supervised pretraining has succeeded in language and vision, and more recently in general time series~\cite{nie2023patchtst,goswami2024moment,das2024timesfm}.

Two obstacles separate that promise from industrial practice. The first is heterogeneity. A PHM corpus mixes vibration sampled at tens of kilohertz with per-cycle counters, one channel with three hundred, and run-to-failure trajectories with steady-state logs; target semantics, in addition, are not comparable across benchmarks. A model trained dataset by dataset does not scale to that diversity; the question is whether a \emph{single} temporal representation can be learned across it and reused. The second obstacle is governance. Condition data belong to different operators and frequently cannot be centralized for reasons of confidentiality, ownership or regulation. Federated learning addresses this in principle~\cite{mcmahan2017fedavg}, but its combination with multi-dataset self-supervised pretraining for PHM has, to our knowledge, not been evaluated empirically under a controlled protocol.

There is also a measurement problem specific to this literature. Recent work has documented that time-series foundation model evaluations are vulnerable to information leakage, both through reuse of datasets across pretraining and testing and through overlapping windows split at random~\cite{meyer2025leakage}. When windows are generated with stride smaller than their length---the norm in PHM---a random split places nearly identical samples on both sides of the train/test boundary and inflates results. Any claim about transfer therefore depends as much on the evaluation protocol as on the model.

This paper is an empirical study, not an architecture proposal. We deliberately fix a small, standard backbone and vary only what we want to measure. Our contributions are:

\begin{enumerate}[leftmargin=*,noitemsep]
  \item \textbf{A leakage-controlled multi-domain PHM corpus and protocol.} We download 53 public datasets, audit 51, harmonize 47 to a common channel-independent tensor contract, and partition them so that 11 datasets are withheld \emph{entirely} from pretraining. Splits are made by physical unit before windowing, normalization is per instance, and automated anti-leakage and round-trip checks pass on 47 of 47 datasets (Section~\ref{sec:corpus}).
  \item \textbf{A budget-matched comparison of centralized and federated self-supervised pretraining.} Both regimes share architecture, objective, source data and optimization budget; only the optimization regime differs. This isolates the cost of not centralizing (Sections~\ref{sec:model} and~\ref{sec:results}).
  \item \textbf{Three empirical findings, reported with their limits.} Transfer is task-dependent and, on RUL, absent; federated pretraining retains part of the representation but does not match the centralized encoder in classification; and the intrinsic pretraining loss does not predict downstream utility, which invalidates it as a model-selection signal. We also quantify how strongly the adaptation learning rate governs whether transfer survives at all.
\end{enumerate}

We consider the negative results to be the more useful half of this paper. Published evidence on when multi-dataset self-supervised pretraining \emph{fails} to transfer in industrial settings is thin, and the failure modes we observe---task dependence, metric decoupling, and destruction of the representation by an overly aggressive fine-tuning rate---are specific, reproducible and actionable.

%% ===== S03_related: Section 2 Related work =====
\section{Related work}\label{sec:related}

\paragraph{Deep learning for PHM.}
Surveys agree on both the capability and the fragmentation of the
field~\cite{zhang2019review,fink2020potential,huang2024phm,ferreira2022rul}. Architectures,
preprocessing and splits are typically designed per task and evaluated on a single benchmark
with intra-dataset protocols. This produces strong specialists and weak evidence about reusable
representations: it is difficult to tell whether a reported gain reflects a transferable inductive
bias or specialization to one controlled environment. The problem of building systems that
transfer across machines, domains and sensor configurations remains open.

\paragraph{Time-series foundation models.}
PatchTST~\cite{nie2023patchtst} introduced two design choices that matter here:
\emph{patching}, which reduces the effective sequence length and provides a more semantic token
than an individual sample, and \emph{channel independence}, which applies a shared encoder to
each channel separately so that no layer depends on the number of sensors.
MOMENT~\cite{goswami2024moment} showed that masked patch prediction over a large multi-dataset
corpus yields a general-purpose encoder evaluated under limited supervision.
TimesFM~\cite{das2024timesfm} and Time-LLM~\cite{jin2024timellm} broaden the design space
towards forecasting-oriented decoders and reprogrammed language models. Surveys organize the
field along modality, backbone, pretraining objective and adaptation
mechanism~\cite{liang2024survey,kottapalli2025survey}. None of these corpora or benchmarks is
built for PHM, and none addresses federated pretraining; we borrow their mechanisms rather than
proposing new ones.

\paragraph{Federated learning under heterogeneity.}
FedAvg~\cite{mcmahan2017fedavg} averages locally trained weights and degrades under non-IID
client distributions through client drift. FedProx~\cite{li2020fedprox} adds a proximal term
that penalizes divergence from the global model; SCAFFOLD~\cite{karimireddy2020scaffold}
corrects drift with control variates; server-side momentum (FedAvgM)~\cite{hsu2019fedavgm}
smooths the aggregation trajectory. Applications to PHM remain narrow in scope, often
partitioning a single dataset artificially, which understates the heterogeneity of a realistic
cross-silo setting where each silo is a distinct physical domain~\cite{zhuang2023fmfl,ren2024ffm}.

\paragraph{Evaluation and leakage.}
TSFM-Bench~\cite{li2024tsfmbench} argues for standardized splits and adaptation protocols.
Meyer et al.~\cite{meyer2025leakage} document leakage in TSFM evaluation, both from dataset
reuse across phases and from temporally overlapping splits. Our protocol targets both failure
modes, and Section~\ref{sec:corpus} reports the outcome of the automated checks.

\paragraph{Positioning.}
The gap we address is the intersection: a temporal self-supervised encoder trained over a
genuinely heterogeneous PHM corpus, compared across centralized and federated regimes, and
evaluated on datasets withheld in full under a protocol that controls leakage and calibrates
adaptation. We do not claim a new architecture, nor state-of-the-art results per benchmark; we
claim a controlled measurement of when this recipe works and when it does not.

%% ===== S04_corpus: Section 3 Corpus and harmonization (3.1-3.3, Table 1, Figures 1 and 2) =====
\section{Corpus and harmonization}\label{sec:corpus}

\subsection{Corpus construction and partition}\label{sec:corpus-construction}

We start from PHMD~\cite{solismartin2025phmd}, a library that unifies access to public PHM datasets. Of 55 candidate
time-series datasets, 53 were downloaded (59.77\,GiB compressed); 51 could be audited (two
failed to load because of external library defects) and 47 satisfied the structural contract. Four
datasets were dropped by explicit, pre-registered criteria---median trajectory length below the
window size, fewer than 100 windows, or tabular behaviour with extreme padding---and none
was dropped because of downstream performance.

\begin{figure}[t]
  \centering
  \includegraphics[width=0.86\textwidth]{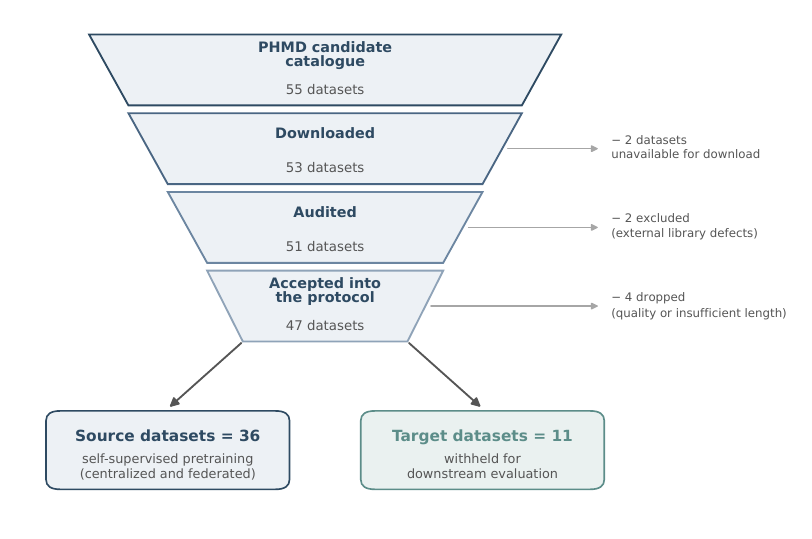}
  \caption{From the PHMD catalogue to experimental roles. Of 55 candidate datasets, 47 satisfy the
  structural contract and are partitioned into 36 source datasets (pretraining only) and 11 target datasets
  (evaluation only). No dataset was dropped on the basis of downstream performance.}
  \label{fig:funnel}
\end{figure}

\begin{figure}[t]
  \centering
  \includegraphics[width=\textwidth]{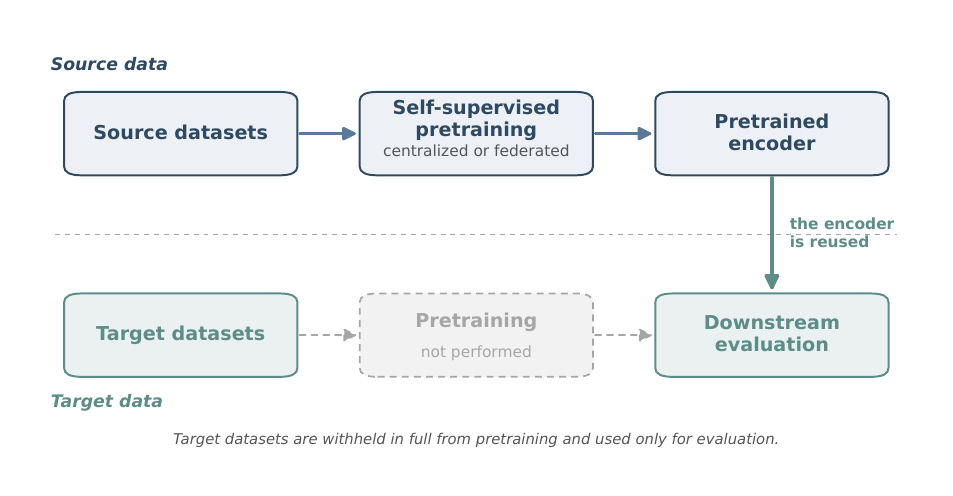}
  \caption{Separation between source and target data. Pretraining, centralized or federated, sees only
  source datasets; the resulting encoder is then adapted and evaluated on target datasets that never entered
  pretraining. The greyed-out path is the one the protocol forbids.}
  \label{fig:source-target}
\end{figure}

The 47 accepted datasets are partitioned into \textbf{36 source} datasets, used \emph{only} for self-supervised
pretraining, and \textbf{11 target} datasets, withheld in full and used \emph{only} for downstream evaluation
(Figure~\ref{fig:funnel}). This partition underpins every claim in the paper (Figure~\ref{fig:source-target}): a target dataset never
participates in pretraining under either regime, so a downstream gain cannot be attributed to
having already seen the domain.

% Table 1 is placed before Figure 2 on purpose: the PDF has Figure 1 (p. 4), Table 1 (p. 5), Figure 2 (p. 6),
% and LaTeX places deferred [t] floats in source order.
\begin{table}[t]
  \centering\small
  \caption{Harmonized corpus. Source totals are the quantities that enter pretraining; the target subset is
  reserved for downstream evaluation only.}
  \label{tab:corpus}
  \begin{tabular}{lr}
    \toprule
    \textbf{Quantity} & \textbf{Value} \\
    \midrule
    Candidate datasets in catalogue & 55 \\
    Downloaded & 53 \\
    Audited & 51 \\
    Accepted into the protocol & 47 \\
    \quad Source (pretraining only) & 36 \\
    \quad Target (evaluation only) & 11 \\
    Dropped / excluded & 4 / 2 \\
    \midrule
    Total windows & 5,686,667 \\
    Total channel patches & 938,903,623 \\
    Processed size & 158.97\,GB \\
    Source valid temporal patches & 175,544,903 \\
    Source valid channel patches & 922,665,279 \\
    \midrule
    Anti-leakage checks passed & 47 / 47 \\
    Round-trip read/write checks passed & 47 / 47 \\
    \bottomrule
  \end{tabular}
\end{table}

The audit inspects the downloaded data rather than catalogue metadata, and records per
dataset: shape and unit structure, missing values and constant channels, target candidates and
coverage, temporal granularity, and post-windowing statistics (windows, valid patches, padding
ratio, estimated size). Aggregate figures for the harmonized corpus are given in Table~\ref{tab:corpus}.

\subsection{Tensor contract}\label{sec:tensor}

Every dataset is mapped to a single interface. A series of a physical unit, $(T, C)$ with $C$ channels, is
segmented into windows of $W = 512$ samples with stride 256; each window is divided into $N = 32$
patches of $P = 16$ samples. Samples are stored as $(C, N, P)$ and batched as $(B, C, N, P)$. Before
the encoder the batch is reshaped to $(B{\cdot}C, N, P)$, so the patch embedding $\mathrm{Linear}(P \to \dmodel)$
and every subsequent layer are shared across channels and independent of $C$. This is what allows
one encoder to process a single-channel disk-drive log and a 317-channel industrial record without
changing a parameter.

Two details carry methodological weight. First, \textbf{validity masks}. A window may be partially
filled when a trajectory ends; the tail is padded and two masks are stored, \texttt{valid\_time\_mask} at
sample resolution and \texttt{valid\_patch\_mask} at patch resolution. The pretraining loss is evaluated
only where a patch is both selected for masking and composed of real samples, so padding never
contributes a gradient. Because the last patch of a padded window mixes real and filler samples,
patch-level masking alone is insufficient; the loss combines both masks at sample resolution,
a behaviour covered by dedicated unit tests. Second, \textbf{tail policy}. We retain the final partial
window with padding rather than discarding it. In degradation data the end of a trajectory is
where failure information lives; discarding tails removed 23.2\,\% of the raw rows of one target
dataset (PHM18) in a preliminary comparison.

Normalization is per window and per channel, computed ignoring padding. This is compatible
with the federated setting---it requires no statistics shared across clients---and avoids a leakage
channel, at the cost of discarding absolute amplitude, which can itself be diagnostic in PHM. We
record the per-window mean and standard deviation on disk so that this choice can be revisited
without reprocessing.

\subsection{Splits and leakage control}\label{sec:splits}

Within each dataset, train, validation and test partitions are formed \emph{by physical unit or trajectory,
before windowing}. With $W = 512$ and stride 256 consecutive windows overlap by half, so splitting
after windowing would place near-duplicate samples on both sides of the boundary. Where a
benchmark provides an official split we respect it, and we construct a global unit identifier from
dataset, original split and unit so that identifier collisions across splits cannot occur. Automated
checks verify that no unit appears in more than one split, that no trajectory is divided, and that
normalization statistics never use test data. These checks pass for all 47 datasets.

%% ===== S05_model: Section 4 Model and pretraining (4.1-4.4, Table 2) =====
\section{Model and pretraining}\label{sec:model}

\subsection{Backbone}\label{sec:backbone}

The encoder, which we refer to as PatchTSTPhm, follows PatchTST~\cite{nie2023patchtst} in its channel-independent
patch formulation and MOMENT~\cite{goswami2024moment} in its pretraining objective. Each patch is projected by a
shared $\mathrm{Linear}(16 \to 128)$, a learnable positional embedding is added, and the sequence passes
through a four-layer Transformer encoder with four attention heads, $\dmodel = 128$ and
pre-normalization. A linear head reconstructs patches during pretraining and is replaced by a task
head downstream. The model has 801,808 trainable parameters.

The model is deliberately small, for three reasons. It is trainable within the compute budget
available to this study; it is cheap to transmit once per federated round, which keeps the federated
arm feasible; and, being fixed and compact, it isolates the variable under study---the origin of the
weights---from capacity effects. We note explicitly that a 0.8M-parameter encoder is two to three
orders of magnitude smaller than published time-series foundation models, and we therefore
make no claim about scale.

\subsection{Self-supervised objective}\label{sec:ssl}

Pretraining uses masked patch prediction. For each window, 30\,\% of the \emph{valid} patches are
replaced by a learnable mask token and the model reconstructs their content from the visible
context; padding is never masked and never scored. The objective requires no labels, which is
what makes the 36 source datasets usable despite their incompatible target semantics.

\subsection{Centralized regime}\label{sec:central}

Sampling is weighted by channel patches, capped so that no dataset exceeds 0.10 and no client
group exceeds 0.25 of the mixture, with a floor of 0.005 to guarantee that the smallest group
participates. The full run executed 99,961 optimizer steps out of 100,000 requested (39 steps
skipped by the AMP gradient scaler on non-finite gradients, a rate of 0.04\,\%) in 3.58\,h on a single
A100. Training loss fell from 0.7974 to 0.4167, a reduction of 47.7\,\%, and the empirical dataset
mixture matched the planned caps to within 0.0018. All 36 source datasets were seen.

\emph{Central-25k}, the budget-matched comparator for the federated arm, is the step-25,000 snapshot
of this same run rather than an independent training run. This has a consequence we return to in
Section~\ref{sec:threats}: its cosine schedule was dimensioned for 100,000 steps, so at step 25,000 the checkpoint
sits in the high-learning-rate phase of a schedule the federated arm does not have. The federated
arm uses a constant learning rate of \sci{3}{-4} without a scheduler, a choice made to isolate the
effect of server momentum. The two arms therefore match in architecture, objective, source data
and number of local optimizer steps, but \emph{not} in optimizer schedule, and the mismatch favours
the centralized arm.

\subsection{Federated regime}\label{sec:federated}

The federated arm simulates a cross-silo setting with ten clients formed by physical domain
(bearings, batteries, aero engines, hard disk drives, wind, gearboxes, CNC milling, and two
deliberately heterogeneous groups). This is a simulation: there are no real organizations, no
network latency, no client dropout and no formal privacy mechanism. What it does reproduce is
the property under study, namely genuine non-IID heterogeneity---clients differ in sensors, scales,
lengths and failure semantics, not in a random partition of one dataset.

Each round distributes the global model, each client performs 50 local steps, and the server
aggregates. With 50 rounds and ten clients this amounts to 25,000 local optimizer steps, matching
Central-25k. Aggregation weights inherit the same caps as the centralized sampler, so the two
regimes see statistically equivalent data mixtures. We evaluate FedAvg and FedAvgM with server
momentum $\beta{=}0.7$, both retained for downstream evaluation, and FedProx with $\mu{=}0.01$, which
was screened and excluded for reasons reported in Section~\ref{sec:intrinsic}.

Table~\ref{tab:intrinsic} reports the intrinsic outcome. At matched budget the centralized encoder reaches
less than half the federated validation reconstruction loss. This is expected---the server averages
once per round, each client optimizes a partial distribution, and the caps redistribute weight
away from dominant domains---but, as we show next, it is a poor predictor of what happens
downstream.

\begin{table}[t]
  \centering\small
  \caption{Intrinsic pretraining outcome. SSL validation loss is a diagnostic of the reconstruction objective,
    not evidence of downstream utility. Central-100k was not evaluated under this protocol, and we do not
    estimate it. All values are computed on ten batches drawn from five of the 36 source datasets (three of
    ten clients) and are comparable only among themselves.}
  \label{tab:intrinsic}
  \begin{tabular}{llrr}
    \toprule
    \textbf{Checkpoint} & \textbf{Regime} & \textbf{Budget (steps)} & \textbf{SSL val. loss} \\
    \midrule
    Central-100k         & centralized & 100,000 & ---    \\
    Central-25k          & centralized &  25,000 & 0.3467 \\
    FedAvg               & federated   &  25,000 & 0.7844 \\
    FedAvgM $\beta{=}0.7$ & federated   &  25,000 & 0.7636 \\
    \bottomrule
  \end{tabular}
\end{table}

%% ===== S06_protocol: Section 5 Downstream evaluation protocol =====
\section{Downstream evaluation protocol}\label{sec:protocol}

We evaluate four pretrained checkpoints---the \emph{checkpoint bank}: Central-100k, Central-25k, FedAvg
and FedAvgM $\beta{=}0.7$---on three held-out tasks: fault classification on CWRU (rolling-element
bearings, 2 channels, 140,134 windows, 153 units) and HSG18 (hard disk drives, 1 channel, 38,896
windows, 17 units), measured by macro-F1; and RUL regression on CMAPSS (24 channels,
comprising 21 sensors and 3 operating-regime variables, 11,795/3,131/6,610 train/validation/test
windows, target capped at 125 cycles).

\paragraph{Composition of the evaluation splits.}
Because splits are formed by physical unit, their
class composition is a consequence of the unit structure rather than a design choice, and it is
uneven. We state it explicitly, since it bounds the interpretation of every classification number
in this paper. The CWRU test split contains 3,331 windows (2.4\,\% of the dataset) drawn from
three of the four fault classes: the fourth class has 6,629 windows in training and none in test.
Reported CWRU macro-F1 is therefore an average over the three classes present in test, and
the class distribution of that split is markedly skewed (477/2378/476). The HSG18 test split
contains 4,576 windows and is exactly balanced (2288/2288). We report macro-F1 throughout
for comparability across tasks; on HSG18 it coincides with accuracy and balanced accuracy to
five decimal places, whereas on CWRU the three quantities differ and the per-class breakdown,
given in Appendix~\ref{app:cwru-classes}, is more informative than any single scalar.

We consider four adaptation modes. \emph{From scratch} trains a randomly initialized encoder on
the target data alone. \emph{Linear probing} freezes the encoder and trains a single layer on top, which
measures representation quality directly. \emph{Full fine-tuning} retrains the whole model with a small
backbone step size. For RUL we add a \emph{two-layer MLP head} over a frozen encoder, since a single
linear layer proved insufficient for continuous regression.

Three rules govern every number we report. \textbf{Selection on validation:} all configuration
choices, including the backbone learning rate and the epoch, are made on the validation split;
the test split is read once, for traceability only, and never used to choose. \textbf{Calibrated learning
rate:} the backbone step size is selected on validation \emph{per dataset and per checkpoint} from
$\{\sci{1}{-6}, \sci{3}{-6}, \sci{1}{-5}\}$, because, as Section~\ref{sec:lr} shows, a single global value silently destroys
some checkpoints. \textbf{Three seeds:} principal comparisons are repeated with seeds 42, 43 and 44
and reported as mean $\pm$ sample standard deviation.

One limitation of the protocol conditions the reading of every table that follows: the \emph{from-scratch}
baselines were run with a single seed, whereas the pretrained checkpoints were run with
three. Comparisons between them are therefore indicative of effect size but not of statistical
reliability.

%% ===== S07_results_cls: Section 6 Results and 6.1 Fault classification (Tables 3 and 4) =====
\section{Results}\label{sec:results}

\subsection{Fault classification}\label{sec:classification}

Table~\ref{tab:classification} reports test macro-F1 under full fine-tuning with calibrated learning rates.

\begin{table}[t]
  \centering\small
  \caption{Fault classification, test macro-F1 (higher is better). Pretrained checkpoints: mean $\pm$ sample
  standard deviation over three seeds, full fine-tuning with the backbone learning rate calibrated on
  validation. \emph{From scratch} and \emph{linear probing} rows are single-seed and are reported as reference
  points, not as matched comparators.}
  \label{tab:classification}
  \begin{tabular}{llcc}
    \toprule
    \textbf{Configuration} & \textbf{Backbone LR} & \textbf{CWRU} & \textbf{HSG18} \\
    \midrule
    From scratch (1 seed) & --- & 0.350 & 0.569 \\
    Linear probing, Central-100k (1 seed) & frozen & 0.705 & 0.906 \\
    \midrule
    Central-100k & \sci{3}{-6} / \sci{1}{-5} & $0.812 \pm 0.033$ & $0.951 \pm 0.002$ \\
    Central-25k & \sci{3}{-6} / \sci{1}{-5} & $0.734 \pm 0.022$ & $0.911 \pm 0.008$ \\
    FedAvg & \sci{3}{-6} / \sci{1}{-5} & $0.572 \pm 0.119$ & $0.742 \pm 0.048$ \\
    FedAvgM $\beta{=}0.7$ & \sci{1}{-6} / \sci{1}{-5} & $0.532 \pm 0.079$ & $0.756 \pm 0.068$ \\
    \bottomrule
  \end{tabular}
\end{table}

\begin{table}[t]
  \centering\small
  \caption{Budget-matched comparisons (Central-25k versus federated), test macro-F1, Welch's $t$-test.
  The direction favours the centralized encoder in all four cases. Two are nominally significant, but none
  survives a Bonferroni correction for four comparisons ($\alpha = 0.0125$). The dispersion being tested comes
  from downstream adaptation seeds over a single pretraining run per arm (Section~\ref{sec:threats}), so these
  $p$-values quantify adaptation noise, not the regime.}
  \label{tab:welch}
  \begin{tabular}{llcccc}
    \toprule
    \textbf{Dataset} & \textbf{Comparison} & $\Delta$ & $t$ & \textbf{df} & $p$ \\
    \midrule
    CWRU & Central-25k $-$ FedAvg & $+0.162$ & 2.31 & 2.14 & 0.139 \\
    CWRU & Central-25k $-$ FedAvgM & $+0.202$ & 4.25 & 2.30 & 0.040 \\
    HSG18 & Central-25k $-$ FedAvg & $+0.169$ & 6.01 & 2.11 & 0.023 \\
    HSG18 & Central-25k $-$ FedAvgM & $+0.155$ & 3.92 & 2.06 & 0.057 \\
    \bottomrule
  \end{tabular}
\end{table}

\paragraph{Pretraining transfers to fault classification.}
On CWRU the centralized encoder more than doubles the from-scratch reference (0.350 to 0.812); on
HSG18 it moves from 0.569 to 0.951. Linear probing alone---an encoder frozen entirely, with a single
trainable layer---already reaches 0.705 and 0.906. That linear probing captures most of the gain
indicates that the improvement comes from the pretrained representation rather than from the
additional optimization afforded by fine-tuning. The effect reproduces across two physically
unrelated domains, rolling-element bearings and disk-drive telemetry, which suggests that the encoder
captures structure common to industrial signals rather than a domain-specific shortcut.

\paragraph{Federated pretraining does not close the gap.}
The budget-matched comparison is Central-25k against the better federated checkpoint. The
centralized encoder leads by 16.2 points on CWRU and 15.5 on HSG18, and it leads in all four pairwise
comparisons (two datasets $\times$ two federated variants). The direction is therefore consistent across
all four comparisons. Statistical reliability is another matter, and we report it plainly in
Table~\ref{tab:welch}. Two of the four comparisons are nominally significant and none survives
correction for four tests. More importantly, the quantity being tested is not the one the claim needs:
our three seeds vary the \emph{downstream adaptation}, while each regime is represented by a single
pretraining run, so the test measures adaptation noise rather than variability of the regime itself
(Section~\ref{sec:threats}). We therefore state the classification result as a large and consistent
effect, not as a statistically established one.

\paragraph{The CWRU ordering is not robust to the learning-rate rule.}
Our protocol selects the backbone learning rate on validation, and on CWRU that rule decides between
two values whose test consequences differ enormously. Central-25k attains validation 0.7794 at
\sci{3}{-6} and 0.7716 at \sci{1}{-5}---a difference of 0.008, well inside seed noise---while the
corresponding test scores are 0.7535 and 0.5551. FedAvg, at those same two rates, tests at 0.6749 and
0.7152. Under the validation rule we obtain the gap reported above; under a single global
\sci{1}{-5}---the value validation selects for every checkpoint on HSG18---the CWRU ordering would
reverse, with FedAvg ahead by roughly 16 points. On HSG18 no such fragility appears: the centralized
encoder leads at all three grid values. The classification claim therefore rests on HSG18, where it
is stable across the grid, and must be read as protocol-dependent on CWRU.

That instability is itself informative. A standard deviation of 0.119 on CWRU test macro-F1,
against 0.022 for the centralized checkpoint at the same budget, indicates that averaging
models trained on strongly non-IID clients produces representations whose downstream behaviour
depends heavily on the adaptation seed.

\paragraph{Budget matters, but does not explain the gap.}
Central-100k outperforms Central-25k on both datasets (0.812 versus 0.734; 0.951 versus 0.911), so
additional pretraining continues to help. Crucially, the federated deficit persists against the
budget-matched Central-25k, so it cannot be attributed to compute alone.

%% ===== S08_results_rul: Subsection 6.2 Remaining useful life (Tables 5 and 6) =====
\subsection{Remaining useful life}\label{sec:rul}

Results on CMAPSS are markedly different. Table~\ref{tab:rul} reports them in full.

\begin{table}[t]
  \centering\small
  \caption{CMAPSS RUL, RMSE in cycles (lower is better), mean $\pm$ sample standard deviation over three
    seeds. The from-scratch reference is single-seed and uses a linear head, so it is \emph{not} a matched comparator;
    it is shown because it bounds the practical value of pretraining on this task.}
  \label{tab:rul}
  \begin{tabular}{llcc}
    \toprule
    \textbf{Checkpoint} & \textbf{Adaptation} & \textbf{Validation} & \textbf{Test} \\
    \midrule
    From scratch (1 seed) & linear head, full training & 39.08 & \textbf{43.67} \\
    Central-100k (1 seed) & linear probing & 43.21 & 53.86 \\
    Central-100k (1 seed) & full fine-tuning & 39.54 & 44.33 \\
    \midrule
    Central-100k & MLP head, frozen & $38.84 \pm 0.07$ & $44.16 \pm 0.21$ \\
    Central-25k & MLP head, frozen & $39.07 \pm 0.06$ & $44.44 \pm 0.58$ \\
    FedAvg & MLP head, frozen & $40.39 \pm 0.06$ & $42.72 \pm 0.15$ \\
    FedAvgM $\beta{=}0.7$ & MLP head, frozen & $38.50 \pm 0.06$ & $44.04 \pm 1.02$ \\
    \midrule
    Central-100k & full fine-tuning & $39.53 \pm 0.28$ & $44.51 \pm 1.13$ \\
    Central-25k & full fine-tuning & $39.85 \pm 0.27$ & $43.83 \pm 0.71$ \\
    FedAvg & full fine-tuning & $39.39 \pm 0.04$ & $44.79 \pm 0.20$ \\
    FedAvgM $\beta{=}0.7$ & full fine-tuning & $39.63 \pm 0.06$ & $45.54 \pm 1.30$ \\
    \bottomrule
  \end{tabular}
\end{table}

\paragraph{No reliable transfer.}
Under the protocol's own rule---select on validation---the best checkpoint
is FedAvgM with 38.50, and its test RMSE is 44.04, statistically indistinguishable from Central-100k's
44.16 ($t{=}0.20$, df$\approx$2.2). Meanwhile the from-scratch reference attains 43.67 on test, better
than three of the four pretrained checkpoints under the MLP head. Validation and test therefore
order the models differently, and the ordering that the protocol commits to does not produce a
model that beats training from scratch.

\paragraph{Why the comparison is even weaker than it looks.}
The from-scratch run uses a linear
output head and trains the full network for a single seed; the multi-seed rows use a two-layer
MLP head over a frozen encoder. Head capacity is therefore confounded with the pretraining
variable, and no matched from-scratch baseline with the MLP head exists. Table~\ref{tab:rul} therefore
does not show that federated pretraining wins on RUL; it shows that this experiment establishes
no benefit of pretraining on CMAPSS.

\paragraph{Under the benchmark's own metric the sign reverses.}
RMSE weighs early and late
predictions equally. CMAPSS ships an asymmetric score that penalizes late predictions---those
that overestimate remaining life, and are therefore the operationally dangerous ones---far more
heavily than early ones. Under that score the single-seed comparison inverts: full fine-tuning of the
centralized encoder scores 645,389 against the from-scratch baseline's 754,754, an improvement
of 14.5\,\%, while its RMSE is 0.65 cycles worse (Table~\ref{tab:cmapss-metrics}). We had computed this metric and
initially omitted it; reporting only RMSE would have concealed the fact that the conclusion on
this task depends on which error model one adopts. The honest summary is that pretraining does
not improve symmetric error on CMAPSS and does appear to reduce the cost of late predictions,
on one seed.

\begin{table}[t]
  \centering\small
  \caption{CMAPSS, single-seed basic modes: the two metrics disagree. The asymmetric score is the
    benchmark's official criterion; lower is better for both columns.}
  \label{tab:cmapss-metrics}
  \begin{tabular}{lccc}
    \toprule
    \textbf{Mode} & \textbf{Test RMSE} & \textbf{Asymmetric score} & $R^2$ \\
    \midrule
    From scratch & \textbf{43.67} & 754,754 & $-0.40$ \\
    Linear probing & 53.86 & 1,002,485 & $-1.13$ \\
    Full fine-tuning & 44.33 & \textbf{645,389} & $-0.44$ \\
    \bottomrule
  \end{tabular}
\end{table}

\paragraph{The test split is structurally different.}
The coefficient of determination on test is negative
for every mode ($R^2 = -0.40$ from scratch, $-0.44$ full fine-tuning, $-1.13$ linear probing), meaning
every model performs worse than predicting the test mean. The cause is a distribution shift
intrinsic to the benchmark: CMAPSS test trajectories are truncated before failure, so the test
target distribution differs from the training one. Absolute prognostic quality is consequently
not interpretable here, and only relative comparisons between models evaluated under the same
protocol carry information. Relative comparisons under a fixed protocol remain informative, so
the numbers are reported with that restriction in mind.

\paragraph{Adaptation mode matters more than provenance.}
Linear probing is by far the worst
mode on RUL (53.86 test RMSE against 43.67 from scratch). A frozen embedding projected by
a single linear layer is not sufficient for continuous degradation regression, whereas the same
frozen embedding with a two-layer head reaches the 42.7--44.4 band. The practical lesson is that
conclusions about ``whether a foundation model transfers'' are contingent on the adaptation head,
and that reporting a single mode invites the wrong conclusion.

%% ===== S09_results_intrinsic: Subsection 6.3 The intrinsic loss does not predict downstream utility (Table 7, Figure 3) =====
\subsection{The intrinsic loss does not predict downstream utility}\label{sec:intrinsic}

The evidence here is weaker than the two preceding subsections and we present it as such.

\paragraph{A large intrinsic gap that does not map onto downstream utility.}
At matched budget the centralized encoder more than halves the federated reconstruction loss (0.3467
versus 0.7844, Table~\ref{tab:intrinsic}). If the intrinsic loss were a good proxy for representation
quality, a factor of 2.3 should be visible downstream. It is not: on CMAPSS the two regimes are
indistinguishable (Table~\ref{tab:rul}), and on CWRU the ordering depends on the adaptation learning
rate (Section~\ref{sec:lr}). A monotone relation between the two metrics is therefore not supported by
our data.

\begin{table}[t]
  \centering\small
  \caption{Server-momentum sweep for FedAvgM (single seed, selected on validation). The intrinsic
  reconstruction loss is nearly constant while downstream utility varies by an order of magnitude more.}
  \label{tab:momentum}
  \begin{tabular}{lccccc}
    \toprule
    & $\beta{=}0.0$ & $\beta{=}0.3$ & $\beta{=}0.5$ & $\beta{=}0.7$ & $\beta{=}0.9$ \\
    \midrule
    SSL validation loss ($\downarrow$) & 0.784 & 0.778 & 0.765 & \textbf{0.764} & 0.768 \\
    CWRU, linear probing ($\uparrow$) & 0.163 & 0.160 & 0.217 & \textbf{0.371} & 0.273 \\
    HSG18, full fine-tuning ($\uparrow$) & 0.713 & 0.464 & \textbf{0.750} & 0.724 & 0.462 \\
    \bottomrule
  \end{tabular}
\end{table}

\begin{figure}[t]
  \centering
  \includegraphics[width=\textwidth]{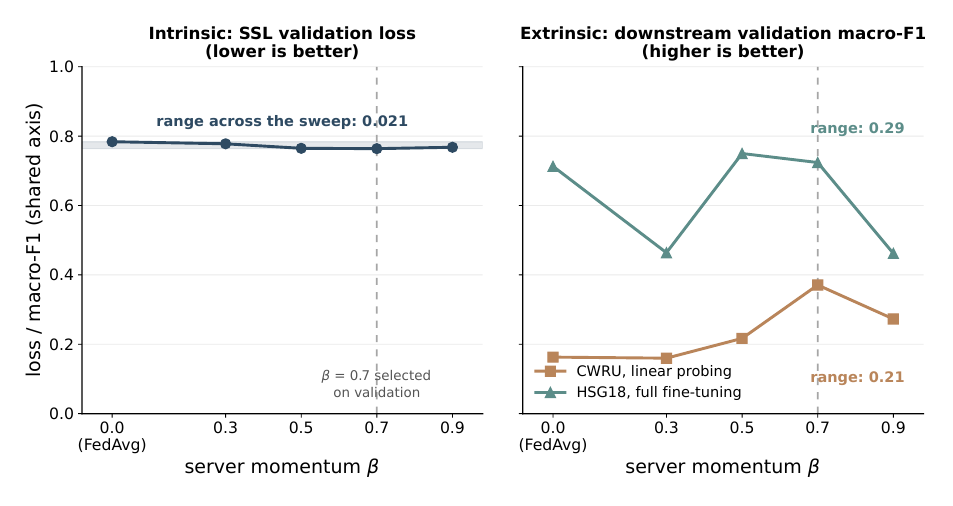}
  \caption{The intrinsic metric is flat where the extrinsic one is not. Server-momentum sweep for
  FedAvgM (data of Table~\ref{tab:momentum}; single seed, screening). Both panels share the vertical axis.
  The SSL validation loss moves within 0.021 across the sweep, while downstream validation macro-F1 moves
  by 0.21 (CWRU, linear probing) and 0.29 (HSG18, full fine-tuning), including the collapse at
  $\beta{=}0.9$.}
  \label{fig:momentum}
\end{figure}

\paragraph{A counter-example to our own claim: FedProx.}
It would be convenient to add FedProx here, and the thesis version of this work did. In a matched
5,000-step screening run FedProx reduced the aggregated reconstruction loss by 11.57\,\% against
FedAvg's 7.55\,\% (0.7423 versus 0.7670). Its downstream behaviour in that same screening, however, was
\emph{also} mostly better than FedAvg's---it won on HSG18 linear probing by a wide margin and lost
clearly only on CWRU fine-tuning. FedProx was excluded from the final bank because its advantage did
not persist when the budget was scaled to 25,000 steps, where FedAvgM dominated it, not because its
pretraining loss misled us. At the screening budget, FedProx is evidence \emph{for} a correspondence
between intrinsic and extrinsic quality, not against it. We report this because it is the clearest case
in our study where a convenient narrative did not survive checking the underlying runs.

\paragraph{Server-momentum sweep.}
Table~\ref{tab:momentum} and Figure~\ref{fig:momentum} sweep the FedAvgM server momentum. The SSL
validation loss varies within a band of 0.021 across the whole sweep---essentially flat---while
downstream validation macro-F1 varies by 0.21 on CWRU and by 0.29 on HSG18, including a collapse at
$\beta{=}0.9$. The two quantities are not merely imperfectly correlated; over this sweep the intrinsic
metric carries almost no information about the extrinsic one.

The consequence for practice is concrete: a time-series foundation model cannot be validated on its
pretraining curves. Checkpoint selection requires evaluation on target tasks, with selection performed
on a validation split.

%% ===== S10_lr_discussion: Subsection 6.4 and Section 7 Discussion (Figure 4) =====
\begin{figure}[t]
  \centering
  \includegraphics[width=0.91\textwidth]{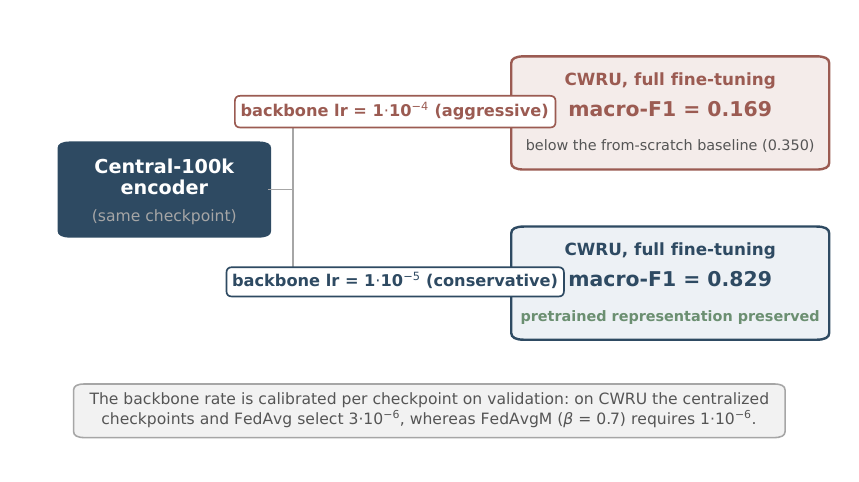}
  \caption{The same checkpoint, two outcomes. Adapting Central-100k to CWRU with a backbone
  learning rate of \sci{1}{-4} overwrites the pretrained representation and lands below the from-scratch baseline;
  \sci{1}{-5} preserves it. Both runs use a single seed; the rates selected on validation for each checkpoint are
  given in the text.}
  \label{fig:lr}
\end{figure}

\subsection{The backbone learning rate governs whether transfer survives}\label{sec:lr}

Adapting the \emph{same} centralized checkpoint to CWRU with a backbone step size of \sci{1}{-4} yields
0.169 test macro-F1---worse than training from scratch---whereas \sci{1}{-5} yields 0.829 (Figure~\ref{fig:lr}).
The aggressive rate overwrites the pretrained representation, a catastrophic-forgetting effect in
which fine-tuning is actively harmful.

The optimum is not a global constant. On CWRU, Central-100k, Central-25k and FedAvg
select \sci{3}{-6} on validation, whereas FedAvgM requires \sci{1}{-6}; the validation difference between
FedAvgM's optimum and the canonical \sci{1}{-5} is nearly 0.10 macro-F1. Had we fixed a single
rate across the bank, FedAvgM would have appeared substantially worse than it is, and we
would have reported a conclusion about federated representations that would in fact have been a
conclusion about an uncalibrated hyperparameter. On HSG18 and on CMAPSS full fine-tuning,
by contrast, \sci{1}{-5} is optimal for all four checkpoints.

This is an evaluation-design result rather than a modelling one: comparisons between
pretrained checkpoints are only meaningful when the adaptation rate is calibrated per checkpoint
on validation.

\section{Discussion}\label{sec:discussion}

\paragraph{Why classification and RUL diverge.}
Fault classification on these benchmarks is largely
separable by local morphology---impulses, frequency signatures, transient shapes---which is
precisely the structure masked patch prediction must model in order to reconstruct hidden
patches. RUL is different: it demands a monotone mapping from window-level evidence to a
global degradation coordinate, information that is distributed across a trajectory rather than
contained in a window. A window-level encoder pretrained on reconstruction has little reason to
encode it. We offer this as an interpretation consistent with the evidence, not as a demonstrated
mechanism; testing it would require architectures with explicit trajectory-level aggregation.

\paragraph{What the federated deficit does and does not show.}
At matched budget the federated
encoder retains a substantial part of the representation---clearly above from scratch in
classification---but does not reach the centralized encoder, and it is markedly less stable across
seeds. Both effects are consistent with client drift under strong non-IID conditions: each client
optimizes a partial and physically distinct distribution, the server averages once per round, and
the caps deliberately move weight away from the dominant domains. Our experiment does not
separate these causes; it measures their joint cost, which is the quantity a practitioner deciding
whether to centralize actually faces.

\paragraph{Scale.}
Our encoder has 0.8M parameters. It is possible that the qualitative conclusions change
with two or three additional orders of magnitude---in particular, larger encoders may transfer
to RUL where ours does not. What our study establishes is that, at this scale and under this
protocol, the pattern is task-dependent transfer, an unclosed federated gap and metric decoupling.
We consider the decoupling finding the most likely to persist across scales, since it concerns the
relationship between two metrics rather than the capacity of the model.

%% ===== S11_threats_conclusion: Section 8 Threats to validity and limitations, Section 9 Conclusion, Reproducibility =====

\section{Threats to validity and limitations}\label{sec:threats}

The first four items below are not peripheral caveats: each of them limits a headline claim,
and two of them we consider serious enough that the corresponding result should be treated as
preliminary.

\begin{itemize}[leftmargin=*,noitemsep]
\item \textbf{The regimes differ in optimizer schedule, not only in regime.} Central-25k is a mid-cosine
snapshot of the 100,000-step run, whereas the federated arm uses a constant learning
rate without a scheduler. The mismatch favours the centralized arm, and we cannot separate
schedule from regime. The classification comparison should be read as an upper bound on the
cost of federating, and a matched Central-25k trained with the federated arm's constant rate is
the single experiment most needed to settle it. Its configuration is committed in the repository
(\path{training/configs/ssl_central_25k_constlr_seed*.yaml}) and validated by dry run; it
was not executed for this version.
\item \textbf{One pretraining run per regime.} The three seeds vary downstream adaptation, not
pretraining. The unit of randomization implied by a claim about regimes is the pretraining
run, and there $n{=}1$ for each arm. Consequently the $p$-values in Table~\ref{tab:welch} do not license
inference about the regimes; multi-seed federated pretraining checkpoints exist in our artefacts
and repeating the comparison over them is straightforward future work; the corresponding
evaluation configuration (\path{downstream_paper_v2_regime.yaml}) is committed alongside the
pretraining configuration above.
\item \textbf{Uncalibrated baselines.} Every from-scratch baseline was run with a single backbone learning
rate of \sci{1}{-4} and no grid, while each pretrained checkpoint received a validation-selected
rate from a three-point grid. The baselines are therefore lower bounds rather than matched
comparators, and the reported classification improvements are upper bounds on the true
effect. A learning-rate grid for the baselines is likewise committed (\path{downstream_paper_v2_fromscratch.yaml}) and not executed.
\item \textbf{CMAPSS departs from the corpus contract.} The RUL dataset is rebuilt by a dedicated
pipeline with causal windows and stride 5, not the stride-256 contract used elsewhere, and
its median trajectory (184 cycles) is shorter than the 512-sample window, so every window is
padded. The RUL null may reflect an encoder operating far outside its pretraining regime
rather than an absence of transferable structure; a $W \in \{64, 128\}$ ablation is required before
attributing it to pretraining.
\item \textbf{Statistical power.} Three seeds, with federated standard deviations up to 0.119 (Table~\ref{tab:welch});
the from-scratch baselines are single-seed (Section~\ref{sec:protocol}).
\item \textbf{Class coverage in CWRU.} One of the four fault classes is absent from the test split, which
contains 2.4\,\% of the dataset; reported macro-F1 averages three classes. Our anti-leakage
checks verify unit disjointness but not class coverage, an omission we recommend others avoid.
\item \textbf{Physical window length.} At CWRU's sampling rates, a 512-sample window spans between
roughly 1.3 and 0.3 revolutions of the shaft. Bearing fault signatures that rely on amplitude
modulation at shaft frequency are therefore only marginally observable, which may bound
achievable performance independently of pretraining.
\item \textbf{Evaluation coverage.} Eleven target datasets were reserved; the multi-seed evaluation covers
three. The remaining eight are documented with their status in Appendix~\ref{app:coverage}, including one
negative case (PBCP16) and one dataset excluded from the body because it was run with an
uncapped batch configuration (PHM18).
\item \textbf{No external baselines.} We compare regimes within a fixed architecture, which isolates the
variable of interest but does not situate the results against specialized PHM methods or against
existing time-series foundation models. A linear-probing comparison against a published TSFM
on the same held-out targets is the single most valuable missing experiment.
\item \textbf{Federated simulation.} Ten clients on one machine, with no latency, no partial participation,
no secure aggregation and no differential privacy. The heterogeneity is real; the systems
dimension is not modelled.
\item \textbf{Fixed window in samples.} $W{=}512$ is a computational contract, not a physical duration: it
spans milliseconds in a high-rate vibration channel and hours in a per-cycle counter. Most
PHMD datasets do not expose a recoverable sampling rate, which makes a physically matched
window infeasible without additional metadata.
\item \textbf{Instance normalization.} Per-window normalization discards absolute amplitude, which can
be diagnostic. The statistics are retained on disk so this can be revisited.
\item \textbf{RUL confound.} As discussed, the from-scratch RUL baseline uses a different head from the
multi-seed rows, so head capacity is confounded with the pretraining variable on that task.
\end{itemize}

\section{Conclusion}\label{sec:conclusion}

We measured, under a leakage-controlled protocol on 47 harmonized PHM datasets, what self-supervised
multi-dataset pretraining buys and what federating it costs. Pretraining transfers
strongly to fault classification across two unrelated physical domains, with linear probing alone
capturing most of the gain. It does not transfer reliably to remaining-useful-life estimation on
CMAPSS, where a from-scratch baseline remains competitive and validation and test disagree.
Federated pretraining at matched budget retains part of the representation but does not close
the classification gap, and produces markedly less stable downstream behaviour. Finally, the
self-supervised loss is not a valid criterion for selecting among pretrained checkpoints: a regime
that more than halves the reconstruction loss is indistinguishable from the other on RUL, and a
hyperparameter sweep can leave the intrinsic metric flat while downstream utility varies by an
order of magnitude more.

The practical recommendations that follow are narrow but firm. Withhold evaluation datasets
in full and split by physical unit before windowing. Calibrate the adaptation learning rate per
checkpoint on validation, or risk attributing to the representation what belongs to the optimizer.
Report linear probing alongside fine-tuning, since their divergence is informative. And never
select a pretrained encoder on its pretraining curve.

\section*{Reproducibility}

Code, configurations, corpus construction and audit artefacts are available at
\url{https://github.com/sanzzjose/fm-predictive-maintenance}. Every reported number is traceable to a versioned
run record containing the configuration hash, the seed, the commit and the resulting
metrics. Raw datasets are redistributed by PHMD~\cite{solismartin2025phmd} and are not mirrored in the repository.

This article extends the first author's master's thesis at Universidad de Sevilla (2026). Relative
to the thesis, the present version adds the significance analysis of Table~\ref{tab:welch}, the re-examination of
FedProx in Section~\ref{sec:intrinsic}, and the first three items of Section~\ref{sec:threats}. The three experiments that those
items identify as most needed are specified as committed, dry-run-validated configurations, so
that they can be executed without further design decisions.

%% ===== S12_appendices: Appendices A-D (Tables 8, 9 and 10) =====
\appendix

\section{Coverage of the held-out target datasets}\label{app:coverage}

Eleven datasets were withheld from pretraining. Table~\ref{tab:targets} documents the status of each, so that
the three used in the body can be read against the full reserve rather than as a selection.

\begin{table}[htbp]
  \centering\small
  \caption{Status of the eleven held-out target datasets.}
  \label{tab:targets}
  \begin{tabular}{lll}
    \toprule
    \textbf{Dataset} & \textbf{Status} & \textbf{Reason} \\
    \midrule
    CWRU & final, $n{=}3$ & classification, body of the paper \\
    HSG18 & final, $n{=}3$ & classification, body of the paper \\
    CMAPSS\_RUL & final, $n{=}3$ & RUL, body of the paper \\
    PBCP16 & diagnostic, $n{=}1$ & negative case, Appendix~\ref{app:pbcp16} \\
    PHM18 & excluded from body & run with uncapped batch; pending re-execution \\
    PHMAP23 & documented negative & extreme class imbalance, noise floor \\
    CALCE\_CS2 & blocked & target semantics (RUL/SOH) unresolved \\
    IEEE14 & not evaluated & RUL target; no dedicated pipeline (only CMAPSS) \\
    CBM14 & not evaluated & secondary tier \\
    CNCMILL18 & not evaluated & secondary tier \\
    PHME20 & not evaluated & secondary tier \\
    \bottomrule
  \end{tabular}
\end{table}

\section{A negative case: PBCP16}\label{app:pbcp16}

PBCP16 is small (1,975 windows, 5 balanced classes, one channel). A from-scratch model reaches
0.907 macro-F1; linear probing reaches 0.829 and full fine-tuning 0.889. Pretraining does not
help because there is no headroom: the task is already solved from scratch. We report this
because it bounds the claim. Self-supervised pretraining is valuable where the target task is hard
relative to its available supervision, not universally, and a study that reported only CWRU and
HSG18 would overstate its scope.

\section{Per-class breakdown on CWRU}\label{app:cwru-classes}

Table~\ref{tab:cwru-classes} gives the per-class decomposition of the best CWRU configuration (centralized checkpoint,
full fine-tuning, \sci{1}{-5}, single seed, test macro-F1 0.8292). Two facts are invisible in the scalar.
First, class 3 has no test support, so the macro average is over three classes. Second, class 0
is recalled almost perfectly but with precision 0.480: 477 windows of class 1 are predicted as
class 0. Whether that matters depends on which class encodes the healthy state---an operational
question that a macro-F1 of 0.83 does not answer.

\begin{table}[t]
  \centering\small
  \caption{CWRU per-class performance, best configuration. Confusion matrix rows are true classes,
    columns predicted.}
  \label{tab:cwru-classes}
  \begin{tabular}{lcccc}
    \toprule
    \textbf{Class} & \textbf{Support} & \textbf{Precision} & \textbf{Recall} & \textbf{F1} \\
    \midrule
    0 & 477 & 0.480 & 0.998 & 0.648 \\
    1 & 2378 & 0.997 & 0.799 & 0.887 \\
    2 & 476 & 0.998 & 0.910 & 0.952 \\
    3 & 0 & --- & --- & undefined \\
    \midrule
    \multicolumn{4}{l}{Macro-F1 over the three supported classes} & 0.829 \\
    \bottomrule
  \end{tabular}

  \medskip
  \begin{tabular}{l|cccc}
    \multicolumn{5}{l}{\footnotesize Confusion matrix} \\
    \toprule
    & \textbf{pred.~0} & \textbf{pred.~1} & \textbf{pred.~2} & \textbf{pred.~3} \\
    \midrule
    true 0 & 476 & 0 & 1 & 0 \\
    true 1 & 477 & 1901 & 0 & 0 \\
    true 2 & 38 & 5 & 433 & 0 \\
    true 3 & 0 & 0 & 0 & 0 \\
    \bottomrule
  \end{tabular}
\end{table}

\section{Hyperparameters}\label{app:hyper}

\begin{table}[htbp]
  \centering\small
  \caption{Principal configuration. Identical across regimes unless stated.}
  \label{tab:hyper}
  \begin{tabular}{ll}
    \toprule
    \textbf{Parameter} & \textbf{Value} \\
    \midrule
    Window / stride / patch & 512 / 256 / 16 \\
    Patches per window ($N$) & 32 \\
    Mask ratio & 0.30 of valid patches \\
    \dmodel{} / layers / heads & 128 / 4 / 4 \\
    Normalization & pre-norm; per-window, per-channel instance norm \\
    Trainable parameters & 801,808 \\
    Sampling caps (dataset / client / floor) & 0.10 / 0.25 / 0.005 \\
    Central budget & 100,000 steps (full), 25,000 (matched) \\
    Federated budget & 50 rounds $\times$ 50 local steps $\times$ 10 clients \\
    Federated algorithms & FedAvg, FedAvgM ($\beta{=}0.7$), FedProx ($\mu{=}0.01$) \\
    Downstream head LR & \sci{1}{-3} \\
    Backbone LR grid & $\{\sci{1}{-6}, \sci{3}{-6}, \sci{1}{-5}\}$ \\
    Seeds & 42, 43, 44 \\
    Hardware & single NVIDIA A100 \\
    \bottomrule
  \end{tabular}
\end{table}

%% ===== S13_references: References (thebibliography with 19 entries) and \end{document} =====

\end{document}